\documentclass[review]{elsarticle}

\usepackage{hyperref}
\usepackage{letltxmacro}
\usepackage{mathtools}
\usepackage{amssymb}
\usepackage{graphicx}
\usepackage{booktabs}

\begin{document}

\begin{frontmatter}

\title{An autoencoder wavelet based deep neural network
with attention mechanism for multistep prediction of
plant growth}


\author[1]{Bashar Alhnaity\corref{cor1}}
\cortext[cor1]{Corresponding author}
\ead{alhnaity.bashar@gmail.com}
\author[1]{Stefanos Kollias}
\author[1]{Georgios Leontidis}
\author[1]{Shouyong Jiang}
\author[2]{Bert Schamp}
\author[3]{Simon Pearson}



\address[1]{School of Computer Science, University of Lincoln, Brayford Pool, LN67TS, Lincoln, UK}
\address[2]{School of Engineering, University of Lincoln, Brayford Pool, LN67TS, Lincoln, UK}
\address[3]{Lincoln Institute for Agri-Food Technology, University of Lincoln, Riseholme Park, LN22LG, Lincoln, UK}
\begin{abstract}
Multi-step prediction is considered of major significance for time series analysis
in many real life problems. Existing methods mainly focus on one-step-ahead
forecasting, since multiple step forecasting generally fails due to accumulation
of prediction errors. This paper presents a novel approach for predicting plant
growth in agriculture, focusing on prediction of plant Stem Diameter Variations
(SDV). The proposed approach consists of three main steps. At first, wavelet
decomposition is applied to the original data, as to facilitate model fitting and
reduce noise in them. Then an encoder-decoder framework is developed using
Long Short Term Memory (LSTM) and used for appropriate feature extraction
from the data. Finally, a recurrent neural network including LSTM and an
attention mechanism is proposed for modelling long-term dependencies in the
time series data. Experimental results are presented which illustrate the good
performance of the proposed approach and that it significantly outperforms the
existing models, in terms of error criteria such as RMSE, MAE and MAPE.
\end{abstract}
\begin{keyword}
multistep prediction, wavelet analysis, LSTMs, deep neural
networks, attention mechanism, time series analysis, plant growth prediction
\end{keyword}
\end{frontmatter}

\section{Introduction}
Time-series analysis and prediction has been a research topic of significance
in various fields and real-life applications, including smart agriculture and prediction
of plant growth, forecasting of financial stocks, anomaly, or intrusion,
detection, medical imaging and air pollution prediction \cite{x0, x2, x10}. Time series
data are generally produced as series of observations aggregated in chronological
order. Their complexity is generally quite high, which makes their analysis a
very challenging task. Due to this nature, using shallow machine learning
and neural network models to analyze the data has produced many bottlenecks.
As a consequence, the development and use of more complex models, which can
automatically extract and learn deep representations from time-series, or image
data, has been a topic of major recent work \cite{x1, x28, x23, x3, x34}.
Recently, Deep Learning (DL) models have produced great progress in agricultural
tasks, such as crop management and plant growth analysis. Plants,
like other bio-systems, are highly complex and dynamic systems. Modelling
plant growth dynamics is a unique challenge, due to large data variations, e.g.,
related to scale of interest, level of description, or integration of environmental
parameters \cite{x2}.
Multi-step time series prediction refers to prediction of the time series in
multiple time steps ahead into the future. In comparison with one step ahead
prediction, multi-step ahead prediction can provide additional benefits to growers;
it is, however, more challenging task as it has to address various additional
complications \cite{x30, x32}. In the literature, there are three primary strategies for
managing multi-step ahead prediction tasks: the recursive strategy, the direct
strategy and the multiple output prediction strategy. The recursive strategy is
based on consecutive one-step-ahead forecasts; each step ahead prediction uses
previously predicted values as inputs. Recursive strategy methods have few
drawbacks, such as error accumulation. The direct strategy predicts separate
models for each forecast. Apart from hybrid direct-recursive strategies, there is
the multi-output model strategy, which is designed to forecast
the entire time series ahead, in one shot. All strategies include challenges that need to be tackled \cite{x31}.
This paper proposes a novel deep learning direct strategy approach for effective
prediction of plant growth. It consists of three components: Wavelet Transformation (WT), encoding-decoding based on LSTM model, and prediction using
LSTM with an attention mechanism. WT can assist in smoothing the noise
effect existing in time series data. The encoder-decoder (ED) part can extract
appropriate features from the reconstructed smoothed signal; these features
form a compact representation, exploited in the final prediction step. A model
composed of LSTM units is blended with Attention Mechanism (AM), in order
to implement the final prediction of plant growth. The resulting approach is
named WT-ED-LSTM-AM hereafter. The effectiveness of the WT-ED-LSTMAM
model is validated using real datasets provided by European greenhouses.
Moreover, the obtained results are compared with those achieved when using
Support Vector Regression (SVR), Random Forest Regression (RFR), standard
Long-Short Term Memory (LSTM) networks,multi-layer perceptrons (MLP),
and networks with gated recurrent units (GRU). An ablation study has also
been implemented, by removing either the wavelet transform part (ED-LSTMAM
method), or the attention mechanism part (WT-ED-LSTM method).
In summary, the main contributions of this paper are the following:
\begin{itemize}
    \item A novel architecture for multi-step prediction of plant growth and stem diameter
variations, including wavelet transformation, data encoding-decoding
and an LSTM with attention components.
\item Improved performance in multi-step prediction on real life data sets, when
compared with baseline models and state-of-the-art methods.
\end{itemize}

The organization of the paper is as follows: Section \ref{s2} presents related
work. Section \ref{s3} describes the proposed pipeline and the utilized models and
components. Section \ref{s4} provides a detailed presentation of the proposed WT-ED-
LSTM-AM approach. Section \ref{s5} presents the developed experimental study.
Finally, Section \ref{s6} provides conclusions and suggestions for future work.

\section{Related Work}\label{s2}

This section provides a short description of existing machine learning prediction
models applied to horticulture, and in particular, to plant growth analysis,
which is crucial for smart farming.
Data-driven models (DDM) that are used in signal processing include Machine
Learning (ML) models, such as Generalized Linear Models, Artificial Neural
Networks \cite{x9} and Support Vector Machines \cite{x26}. Those methods have many
desirable characteristics, such as: imposing few restrictions and assumptions;
ability to approximate nonlinear functions; strong predictive capabilities; 
ability to adapt to multivariate system inputs. According to \cite{x24, x29}
machine learning, linear polarisation, wavelet-based filtering, vegetation indices
and regression analysis are the most popular techniques used for analyzing agricultural
data. Deep Learning (DL) has obtained great popularity in the last few
years \cite{x14}. DL involves Deep Neural Networks (DNNs), which can extract hierarchical feature structures and create rich representations of the data. A strong
advantage of DL is feature learning, i.e., automatic feature extraction from raw
data, with features from higher levels of the hierarchy being formed through
composition of lower-level features. Consequently, DL can solve complex
real life problems with high accuracy, provided there is availability of adequately large data-sets describing the problem. Gonzalez-Sanchez et al. \cite{x13}
presented a comparative study of ANN, SVR, M5-prime regression, K-nearest
neighbor classifiers and Multiple Linear Regression for crop yield prediction
in ten crop datasets. In their study, Root Mean Square Error (RMSE), Root
Relative Square Error (RRSE), Normalized Mean Absolute Error (MAE) and
Correlation Factor (R) were used as accuracy metrics to validate the models.
Results showed that M5-Prime regression achieved the lowest errors across the
produced crop yield models. The results of that study ranked the techniques
from best to worst, as follows: M5-Prime, kNN, SVR, ANN, MLR. Another
study by \cite{x6} applied four ML techniques, SVM, Random Forest (RF), Extremely Randomised Trees (ERT) and Deep Learning (DL) to estimate corn 
yield in Iowa State. Comparison of the validation statistics showed that DL
provided the more stable results, overcoming the over-fitting problem. In the
current paper (and in \cite{x2}) we develop a novel deep learning architecture for prediction
of plant growth, using stem diameter variations as a growth indicator.

Stem diameter is considered a parameter of major importance that describes
the growth of plants during vegetative growth stage. The variation of stem diameter
has been widely used to derive proxies for plant water status and, is
therefore used in optimisation strategies for plant-based irrigation scheduling in
a wide range of species. Plant stem diameter variation (SDV) refers to plant
stem periodic shrinkage and recovery movement during day and night. This
periodic variation is related to plant water content and can be used as an indicator
of the plant water content changes. During active vegetative growth
and development, crop plants rely on the carbohydrate gained from photosynthesis
and the translocation of photo-assimilates from the site of synthesis to
sink organs. 

The fundamentals of stem diameter variations have been well
documented in the literature \cite{x33}. It has been documented that SDV is sensitive
to water and nutrient conditions and is closely related to the response of crop
plants to changes of environmental conditions \cite{x19}. Moreover, stem diameter is
a parameter that describes the growth of crop plants under abiotic stress during
vegetative growth stage. Therefore, it is important to generate stem diameter
growth models able to predict the response of SDV to environmental changes and
plant growth under different conditions. 
Many studies emphasize the need to
critically review and improve SDV models for assessment of environmental impact
on crop growth \cite{x16}. SDV daily models have been developed to accurately
predict inter-annual variation in annual growth in balsam fir (Abies balsamea
L). Inclusion of daily data in growth-climate models can improve prediction
of the potential growth response to climate by identifying particular climatic
events that escape to a classical dendroclimatic approach \cite{x11}. However, development
of models that are capable of predicting SDV and plant growth taking
into consideration environmental variables has so far remained limited.

Since horticulture management decisions become data-driven, DL is continuously gaining popularity as one of the most successful techniques to model
obtained data. In this paper, we propose a DL model and a new approach
for multi-step prediction of plant growth using wavelet transformation (WT),
encoder-decoder based on LSTM and RNN-LSTM prediction with an attention
mechanism and we evaluate its performance on real plant growth data.

\section{Problem Definition and Components} \label{s3}

\subsection{Problem definition} \label{problem}

The aim of a model for single step time series prediction is to implement
a mapping from a sequence of input data to a single output
target value,
where the input is generally an $M$-dimensional vector, with the estimated output occurring $k$ steps ahead.
The model is usually estimated through supervised learning with a direct strategy for multi-step prediction, using a collection of training data and respective
labels.

\subsection{Wavelet transform}\label{WT}
The Wavelet transform can be used for data denoising, while handling the
non-stationary nature of the collected time series data. In the following we
use the wavelet transform for representing, decomposing and reconstructing the
original data. Wavelet analysis was firstly introduced by Mallat \cite{x25} and since
then has been used in various domains for signal processing, image recognition, remote sensing data decomposition, time series decomposition,
medical image analysis and medical diagnosis. The Discrete Wavelet Transform decomposes signals into a low frequency approximation set and several
high frequency detailed sets.

Mallat proposed filtering the time series using a pair of high-pass and low-pass filters as an implementation of discrete wavelet transform. There are two types of wavelets, father wavelets $\varphi(t)$ and mother wavelets $\psi (t)$, in the Mallat algorithm. Father wavelets $\varphi(t)$ and mother wavelets $\psi (t)$ integrate to 1 and 0, respectively, which can be formulated as:
\begin{equation}
\int \varphi (t)dt = 1, \int \psi (t)dt =0
\end{equation}
The mother wavelets describe high-frequency parts, while the father wavelets describe low-frequency components of a time series. The mother wavelets and father wavelets in the $j$ level can be formulated as:
\begin{equation}
\varphi _{j,k}(t) = 2^{-\frac{j}{2}}\varphi (2^{-j} - k)
\end{equation}
\begin{equation}
\psi _{j,k}(t)=2^{-\frac{j}{2}}\psi (2^{-j}-k)
\end{equation}
Time series data can be reconstructed by a series of projections on the mother and father wavelets with multilevel analysis indexed by $k\epsilon \left \{ 0,1,2, \right \}$ and by $j\epsilon \left \{ 0,1,2, J\right \}$, where $J$ denotes the number of multi-resolution scales. The orthogonal wavelet series approximation to a time series $x(t)$ is formulated by:
\begin{equation}
x(t)=\sum _{k}s_{J,k}\varphi _{J,k}(t)+\sum _{k}f_{J,k}\psi _{J,k}(t)+\sum _{k}d_{J-1,k}(t)+...+\sum _{k}d_{1,k}\psi _{1,k}(t)
\end{equation}
where the expansion coefficients $s_{J,k}$ and $d_{J,k}$ are given by the projections
\begin{equation}
s_{J,k}=\int \varphi _{J,k}x(t)dt
\end{equation}
\begin{equation}
d_{j,k}=\int \psi _{j,k}x(t)dt
\end{equation}
The multi-scale approximation of time series $x(t)$ is given as:
\begin{equation}
S_{J}(t)=\Sigma _{k}s_{J,k}\varphi _{J,k}(t)
\end{equation}
\begin{equation}
D_{j}(t)=\Sigma _{k}d_{j,k}\psi _{j,k}(t)
\end{equation}
Then, the brief from of orthogonal wavelet series approximation can be denoted by:
\begin{equation}
x(t)=S_{J}(t)+D_{J}(t)+D_{J-1}(t)+.+D_{1}(t)
\end{equation}
where $S_{J}(t)$ is the coarsest approximation of the input time series $x(t)$. The multi-resolution decomposition of $x(t)$ is the sequence of $S_{J}(t),D_{J}(t),D_{J-1}$ $(t),...D_{1}(t)$. 

There are several wavelet families, such as Daubechies (dbN), Coiflets (CoifN) and Symlets (symN). In this paper we use db2 to decompose the original series
into one approximation and two detail sets.

\subsection{Support vector regression}
Support vector regression (SVR) is a classification method that arises from a nonlinear generalization of the Generalized Portrait algorithm developed by Vapnik \cite{x8}. The goal of SVR is to obtain a linear function with \(f(x)=<w,x>+b\) with \(w\epsilon R^{N}\) and \(b\epsilon R\) for given training set \(\left \{(x_{1},y_{1}),...,(x_{m},y_{m})\right \}\), as follows:
\begin{equation}
\begin{split}
minimize 
\frac{1}{2}\left \| w \right \|^{2}+c\sum_{i=1}^{m}(\xi_{i}+\xi_{1}^{*}) \\
subject   
 to\left\{\begin{matrix}
y_{i}-\left \langle w,x_i \right \rangle-b \leq \varepsilon +\xi _{i}\\ 
\left \langle w,x_{i} \right \rangle +b-y_{i} \leq \varepsilon +\xi_{i}^{*} \\ 
\xi_{i},\xi_{i}^{*} \geq 0 
\end{matrix}\right.
\end{split}
\end{equation}
where $\xi_{i}$ and $\xi_{i}^{*}$ are slack variable introduced to deal with infeasible constraints, and $C$ is called the $regularization$ parameter. In most cases, the problem can be solved in its dual formulation:
\begin{equation}
\begin{split}
max_{a,a^{*}}-\frac{1}{2}\sum _{i=1}^{N}\sum_{j=1}^{N}(a_{i}-a_{i}^{*})(a_{j}-a_{j}^{*})K(x_{i}-x_{j})
\\
-\varepsilon \sum _{i=1}^{N}(a_{i}-a_{i}^{*})+\sum _{i=1}^{N}y_{i}(a_{i}-a_{i}^{*})\\
\text{subject to} :\\
\sum _{i=1}^{N}(a_{i}-a_{i}^{*})=0,a_{i},a_{i}^{*}\varepsilon \left [ 0,C \right ]
\end{split}
\end{equation}
where \(K(x_{i},x_{j})\) is known as the kernel function, which allows to project the original data into a higher dimensional space to be linearly separable. Common kernel functions include the linear, radial basis and polynomial
ones. Among these, Radial Basis Function (RBF) provides dimensionality reduction,
restricting the computational load during training and providing improved
generalization capabilities. For these reasons, RBF kernel has been adopted, defined as follows:
\begin{equation}
K(x,x_{i}) = exp\left ( -\frac{1}{\sigma ^{2}}\left \| x-x_{i} \right \|^{2} \right )
\end{equation}
where $x$ and $x_{i}$ are vector in the input space, i.e. vectors of features computed from training or test samples.

\subsection{Random forest regression}

Random forest regression (RFR) belongs to the category of ensemble learning
algorithms \cite{x17}. As a base learner of the ensemble, RFR uses decision trees. The
idea of ensemble learning is that multiple predictors can be more effective in
making predictions over the test data, distinguishing noise from patterns. RFR
constructs independent regression trees, with a bootstrap sample of the training
data being chosen at each regression tree. As a consequence, the regression tree
continuously grows until it reaches the largest possible size. Final prediction is
a weighted average of all regression trees predictions.

\subsection{Multilayer perceptron}
Multilayer Perceptrons (MLP) \cite{x15} have been  the main architecture used for
supervised learning and classification tasks in the past; they consist of multiple
fully connected layers of neurons, with feedforward spread of information. Their
training is performed with the backpropagation algorithm.

\subsection{Long-short term memories}
Long short-term memory (LSTM) is a variation of the recurrent neural
network (RNN) architecture \cite{x18}. They have been able to solve the gradient
vanishing problem in long-term time series analysis. The LSTM structure
contains three modules: the forget gate, the input gate and the output gate.
The forget and input gates control which part of the information should be removed/
reserved to the network; the output gate uses the processed information
to generate the provided output. LSTMs also include a Cell State, which allows
the information to be saved for a long time. In the following, we illustrate the
operation of LSTM units.

Let $i_{t}$ and $\tilde{C}_{t}$ be the values of the input gate and the candidate state of the memory cell at time $t$, respectively. These are computed as follows:

\begin{equation}
i_{t} =\sigma (W_{i}x_{t} + U_{i}h_{t-1}+b_{i})
\label{Eq:18}
\end{equation}
\begin{equation}
\tilde{c}_{t} =\tanh (W_{c}x_{t} + U_{c}h_{t-1}+b_{c})
\end{equation}
Let us denote by
$f_{t}$ and $C_{t}$ the value of the forget gate and the state of the memory cell at time $t$, respectively. These can be calculated by:
\begin{equation}
f_{t} =\sigma(W_{f}x_{t} + U_{f}h_{t-1}+b_{f})
\end{equation}
\begin{equation}
c{t}=i_{t}*\tilde{c}_{t}+f_{t}*C_{t-1}
\end{equation}
Let, also, $o_{t}$ and $h_{t}$ denote the  values of the output gate and the value of the memory cell at time $t$, respectively. These are computed as follows:
\begin{equation}
o_{t}=\sigma(W_{o}x_{t}+U_{o}h_{t-1}+V_{o}C_{t}+b_{o})
\end{equation}
\begin{equation}
h_{t}=o_{t}*\tanh(c_{t})
\label{Eq:23}
\end{equation}
where $x_{t}$ is the input vector to the memory cell at time t: $W_{i}, W_{f},W_{c},W_{o}, U_{c},U_{o}, V_{o}$ are weight matrices; $b_{i},b_{f}, b_{c}, b_{o}$ are bias vectors.

\subsection{Gated recurrent units}
Gated recurrent units (GRU) are simplified LSTMs. They do not include
output gates, thus there is no control over the memory content. They can be
used instead of LSTMs. Further information can be found in \cite{x7}.

\subsection{LSTM Encoder-Decoder Model}

\begin{figure}[tph!]
\includegraphics[totalheight=7cm]{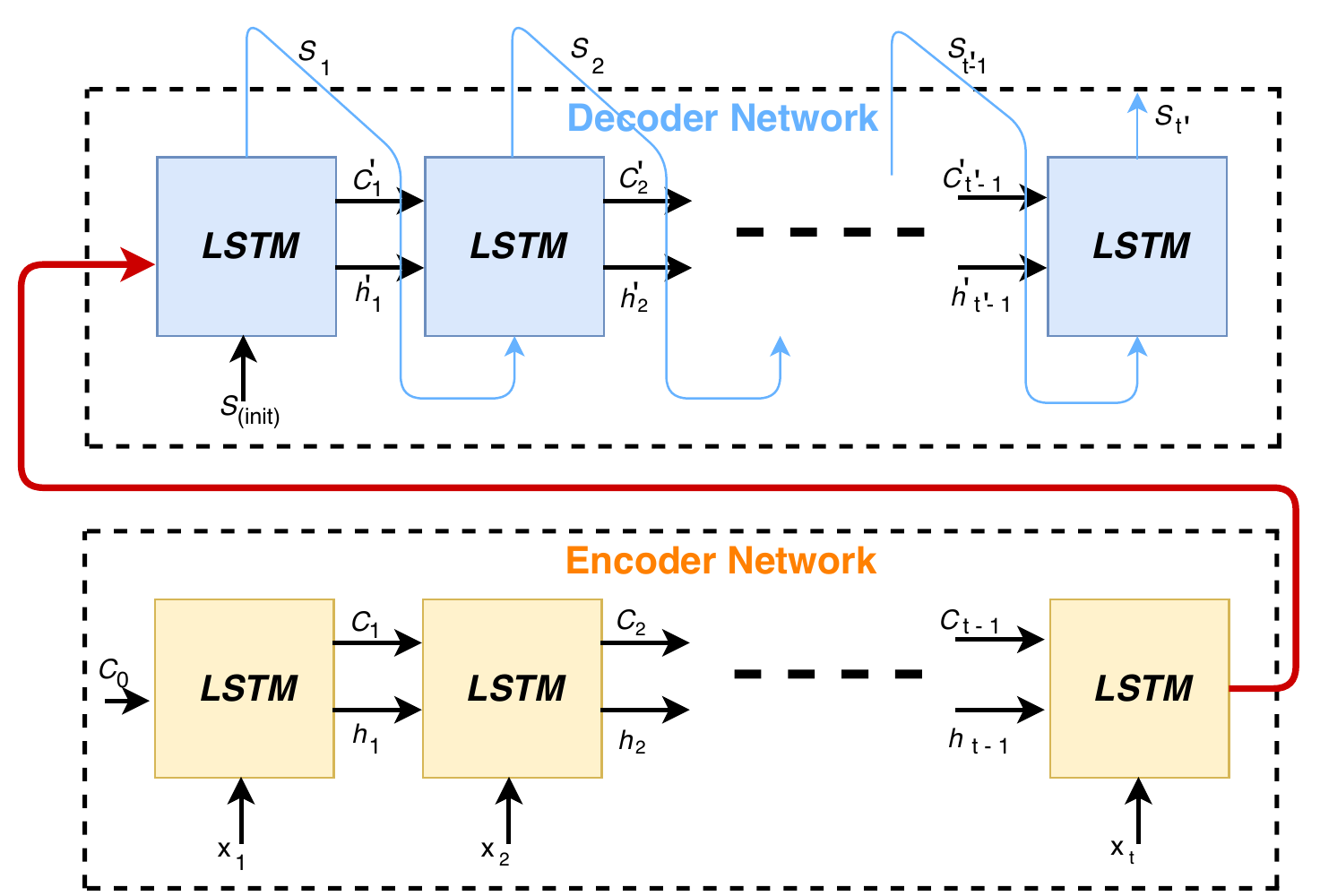}
\centering
\caption{LSTM encoder decoder architecture.}
\label{fig:LSTM_Decoder_Encoder}
\end{figure}

In LSTM Encoder-Decoder models, the encoder part compresses the information
from the entire input sequence into a vector composed of the sequence
of the LSTM hidden states. Consequently, the encoder summarizes the
whole input sequence into the final cell state vector and passes it to the decoder \cite{srivastava2015unsupervised}. The latter uses this representation as initial state to reconstruct
the time series.
 The architecture employs two LSTM networks called the encoder and decoder, as shown in  Fig. \ref{fig:LSTM_Decoder_Encoder}
.

\subsection{Attention Mechanism}

Attention mechanisms have been used as a means to improve performance
in vision and signal processing tasks, by focusing on feature segments of high
significance \cite{x4, x27}. They are currently implemented through attentive neural
network models \cite{x12}. Bahdanau et al \cite{x5} introduced an
attention mechanism to model a long-term dependence, by generating a context
vector as a weighted sum of all provided information. In this paper, the attention
mechanism is used both across the dfferent internal LSTM layers, as well as over
the LSTM output layers. Prediction of the output signal can be derived using
the conditional probability distribution of the input signal and of the previous
sample of the output signal, i.e.,

\begin{equation}
p(y_{i}\mid x_{0},...,x_{n}, y_{i-1})
\label{Eq:22}
\end{equation}

This distribution is, however, impossible to compute  in most real life applications. Equation \ref{Eq:22} can be approximated by the non-linear function:
\begin{equation}
p(y_{i}\mid x_{0},...,x_{n}, y_{i-1})\approx g(y_{i},h_{i},C_{i})
\label{Eq:d22}
\end{equation}
where $g$ is LSTM, $h_{i}$ is the internal state of the LSTM and $C_{i}$ is the current context, i.e., a vector holding information of which inputs are important at the current step. The context is derived from both the current state,$h_{i}$, and the input sequence $x$. 
After the LSTM has stepped through
the whole input sequence, the attention mechanism of the network decides the
attention that should be put on the annotation provided at each step. The
transition functions of the attentive neural network are described by Eqs. \ref{Eq:24}-\ref{Eq:26}.  The attention mechanism begins with computing $e_{t}$ :
\begin{equation}
e_{t}=v^{T} . \tanh \left ( W_{e}. h_{t}+U_{e}.d_{t-1}+b \right )
\label{Eq:24}
\end{equation}
where $v,b,h_{t},d_{t-1}$ $\varepsilon$ $\mathbb{R}^{n}$ and $W_{e}, U_{e}$ $\varepsilon$ $\mathbb{R}^{n*n}$ and $d$ denotes here the input sequence as well. The attention score, $a^{t,t'}$, for each $t'$  is then computed by the softmax function, as follows:
\begin{equation}
a^{t,t'}= \frac{\exp(e_{t}) }{\sum_{t=1}^{T}\exp(e_{t})}
\label{Eq:25}
\end{equation}
The context vector, $C_{t}$,  is computed as the weighted sum of all internal states, $\left \{ h_{1},...,h_{T} \right \}$:
\begin{equation}
C_{t}= \sum_{t'=1}^{T}a^{t,t'}.h_{t'}
\label{Eq:26}
\end{equation}

\section{The Proposed approach} \label{s4}

\subsection{Setting up the prediction framework}
In the following, we use the models and methods, described in the previous
Section, in a novel deep prediction framework. The proposed architecture (WTED-
LSTM-AM) includes wavelet-based transformation of the collected signals,
followed by an encoding-decoding step, using LSTM and attention models for
final prediction.

\begin{figure}[tph!]
\includegraphics[totalheight=10cm]{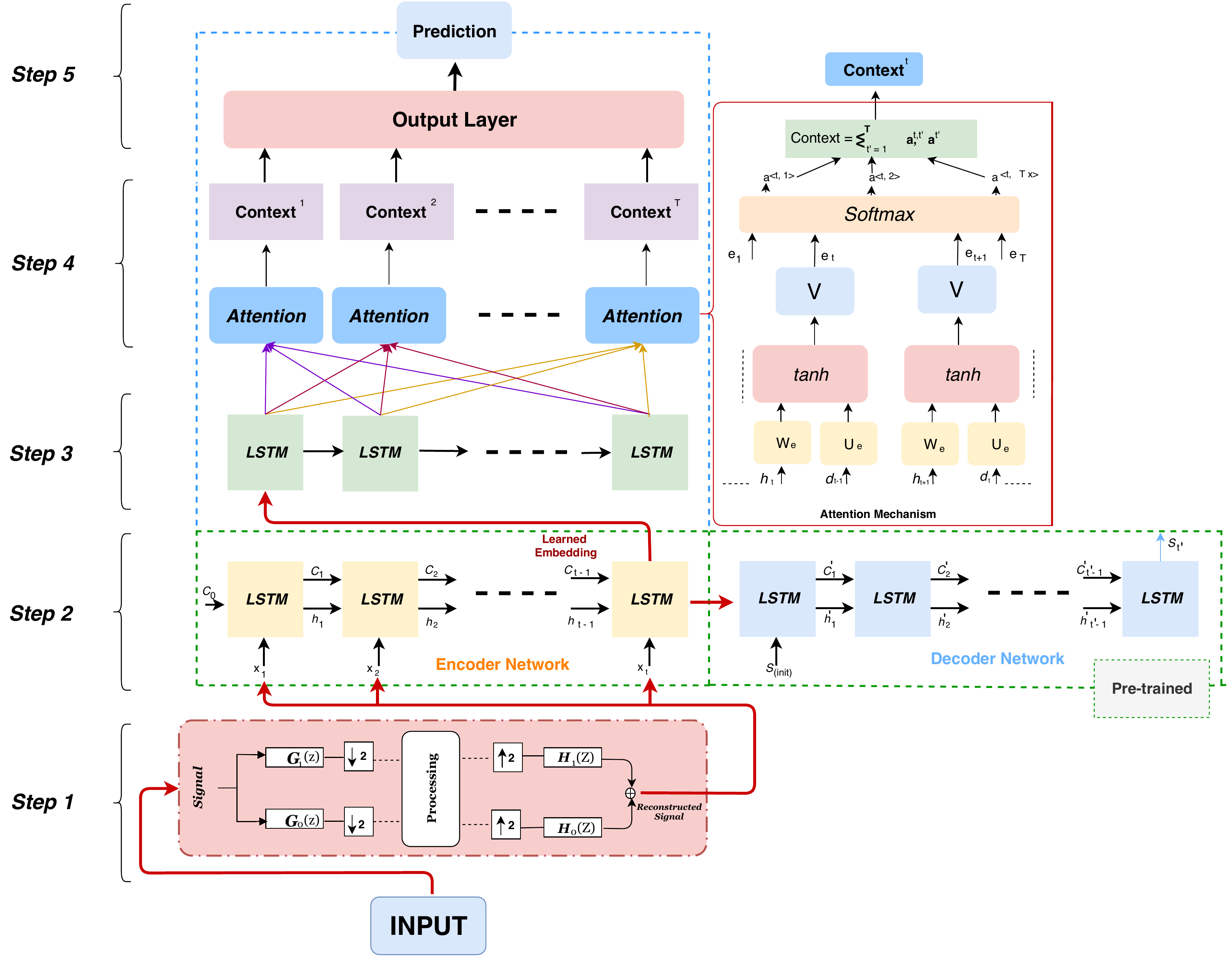}
\centering
\caption{Deep architecture model (WT-ED-LSTM-AM) for SDV prediction.}
\label{fig:proposedapproached}
\end{figure}

Figure \ref{fig:proposedapproached} shows the proposed approach for SDV multi-step prediction. The
target is to predict SDV in multiple hourly steps ahead, based on current information
and history sensory signal data.

\subsection{The WT-ED-LSTM-AM architecture}

The proposed WT-ED-LSTM-AM architecture for plant growth prediction,
shown in Figure 2, is composed of five steps
:
\begin{itemize}
\item Step 1: Data denoising is performed first, using the wavelet transform
(WT). In particular, we decompose each input signal in two components,
generating a subsampled (by 2) time series approximation and eliminating
noise that is present in the high frequency component. By upsampling (by
2) and filtering, a reconstructed signal is obtained, which is provided to
the next step of our approach.

\item Step 2: The encoder-decoder stage is then implemented. The encoder
is pre-trained to extract useful and representative embeddings from the
reconstructed time series data; these embeddings can be usedcfor prediction purposes. Two-layer LSTM cells are used in the encoder
implementation. Based on the learned embedding states, the decoder has
learnt to generate the (reconstructed) input signal. We designed this step,
inspired by the success of video representation learning, where a similar
architecture was introduced \cite{srivastava2015unsupervised}.

\item Step 3: The encoder-decoder step constitutes the feature extraction component
of the proposed approach. Then, an LSTM network, with attention
mechanism, is trained to make single, or multi-step prediction, using the
learned embedding as input features. LSTMs use the transition functions $\left \{h_{1},h_{2},...,h_{n} \right \}$ from the learned embedding states from the previous step. 

\item Step 4: as shown in Fig. \ref{fig:proposedapproached}, the attention mechanism is applied to the
outputs of each LSTM unit to model a respective long-term dependence.
The learned embedding states, the attention weights corresponding to
these states, and the respective context, are computed as described in the
previous Section, are used for implementing the attention mechanism.

\item Step 5: A single layer neural network is responsible for the final prediction
of the SDV value, as described  in Eq. \ref{eq:2520}.
\begin{equation}
h_{s}= tanh(W_{p}C+W_{x}h_{n})
\end{equation}
\begin{equation}
\hat{y}= W_{s}h_{s}+b_{s}
\label{eq:2520}
\end{equation}
\end{itemize}

\section{The Experimental Study} \label{s5}

An extensive experimental study has been carried out to evaluate the performance
of the proposed approach, targeting supervised multi-step prediction
of SDV in real-world data sets. The obtained results illustrate the effectiveness
and efficiency of the proposed approach in predicting the SDV.

\subsection{Experimental set-up}
The proposed architecture was used to predict growth of Ficus plants (Ficus
benjamina), based on data collected from four cultivation tables in a 90 $m^{2}$
greenhouse compartment of the Ornamental Plant Research Centre (PCS) in
Destelbergen, Belgium. Plant density was approximately 15 pots per $m^{2}$, where
every pot contained cuttings.
The experiment started on 23 March 2016. Greenhouse microclimate was
set by controlling the window openings, a thermal screen, an air heating system,
assimilation light and a CO$_2$ adding system. Plants were irrigated with an
automatic 
flood irrigation system, controlled by time and radiation sum. Set-points
for microclimate and irrigation control were similar to the ones used in
commercial greenhouses. The microclimate of the greenhouse was continuously
monitored. Photosynthetic active radiation (PAR) and CO$_2$ concentration were
measured with an LI-190 Quantum Sensor (LI-COR, Lincoln, Nebraska, USA)
and a carbon dioxide probe (Vaisala CARBOCAP GMP343, Vantaa, Finland),
respectively. Temperature and relative humidity were measured with a temperature
and relative humidity probe (Campbell Scientific CS215, Logan, UT,
USA), which was installed in a ventilated radiation shield.

Stem diameter was continuously monitored on three plants with a linear
variable displacement transducer (LVDT, Solartron, Bognor Regis, UK) sensor.
The hourly variation rate of stem diameter ($mm$ d$^{-1}$) was calculated as the
difference between the current stem diameter and the stem diameter recorded
on one hour earlier, at a given time point. Thus, the frequency of collected data
has been at one hour basis.
We performed experiments on one-step, two-step and three-step forecasting.
In one-step-ahead forecasting, we used input data collected in previous 15
hours, to predict the SDV value in the current hour.
In two-step-ahead, i.e., 6 hours forecasting, we used input data collected in
the previous 6 hours, with a 6-hour stride.
In three-step-ahead, i.e., 12 hours forecasting, we used the previous 12 hours,
with a 12-hour stride.

In all experiments, we used the first 70$\%$ of data samples as training set, the
next 10$\%$ of data samples as validation set and the rest 20$\%$ of data samples as
test set.

\subsection{Performance evaluation}
The Mean Absolute Error (MAE), the Root Mean Squared Error (RMSE) and the Mean Squared Error (MSE) were used to evaluate the performance of prediction models. Formulas of these evaluation measures are shown below:
\begin{equation}
MSE = \frac{1}{n}\sum _{t=1}^{n}\left ( \frac{A_{t}-F_{t}}{A_{t}} \right )^{2}
\end{equation}
\begin{equation}
MAE=\frac{1}{n}\sum_{t=1}^{n}\frac{\left | A_{t}-F_{t} \right |}{\left | A_{t} \right |}
\end{equation}
\begin{equation}
RMSE = \sqrt{\frac{1}{n}\sum _{t=1}^{n}\left ( \frac{A_{t}-F_{t}}{A_{t}} \right )^{2}}
\end{equation}
where $A_{t}$ denotes actual values and $F_{t}$ denotes the predicted values.

\subsection{Feature normalization}

In all experiments we used min-max normalization (min-max scaling) on
the extracted features, re-scaling their values in the range [0, 1]. 

\subsection{Experimental results}
\begin{figure}[tph!]
\includegraphics[totalheight=5cm]{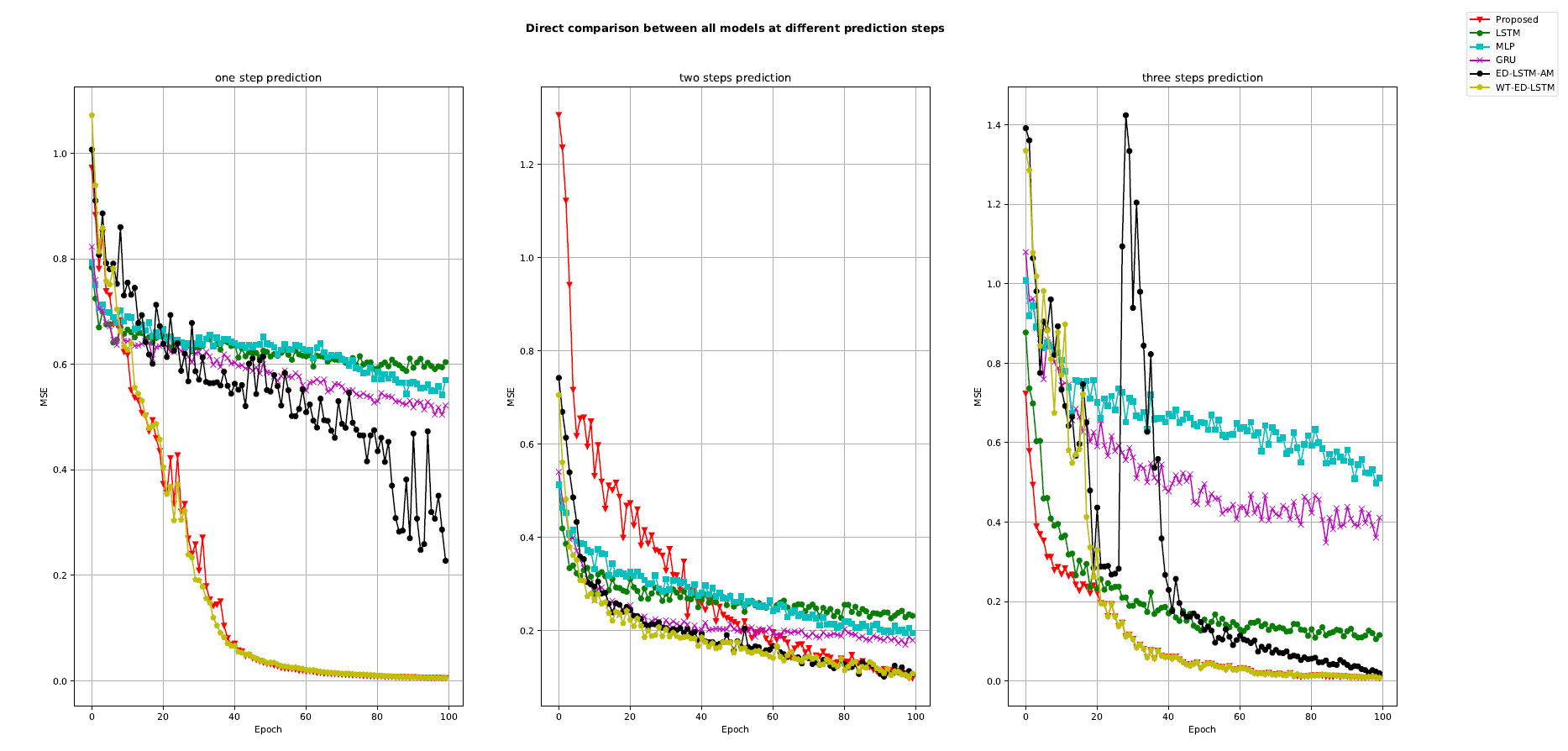}
\centering
\caption{Training comparison of the different models at each epoch.}
\label{fig3}
\end{figure}

The experimental results illustrate the very good performance of the proposed
methodology, which outperforms all considered baseline methods. For
comparison purposes, we used the same hyper-parameters in the proposed approach
and in three baseline models: a two-layer stacked GRU, a LSTM and a MLP with Stochastic Gradient Descent (SGD); learning rate ls = 0:001 and
batch size = 32 were adopted. All models were trained for 100 epochs, using the
same training, as well as validation and test data sets. In the proposed method,
we used a two layer LSTM encoder-decoder structure, with 128 and 32 neurons
respectively. In the prediction model, we used a single layer LSTM with 128
neurons. Figure \ref{fig3} illustrates minimization of the MSE per epoch during the
training phase of all models, in all three prediction tasks.

.

The obtained accuracy in terms of the three error criteria for the multi-step
prediction (1h, 6h, 12h) tasks for all considered methods is shown in the following Table.

\begin{figure}[tph!]
\includegraphics[totalheight=5cm]{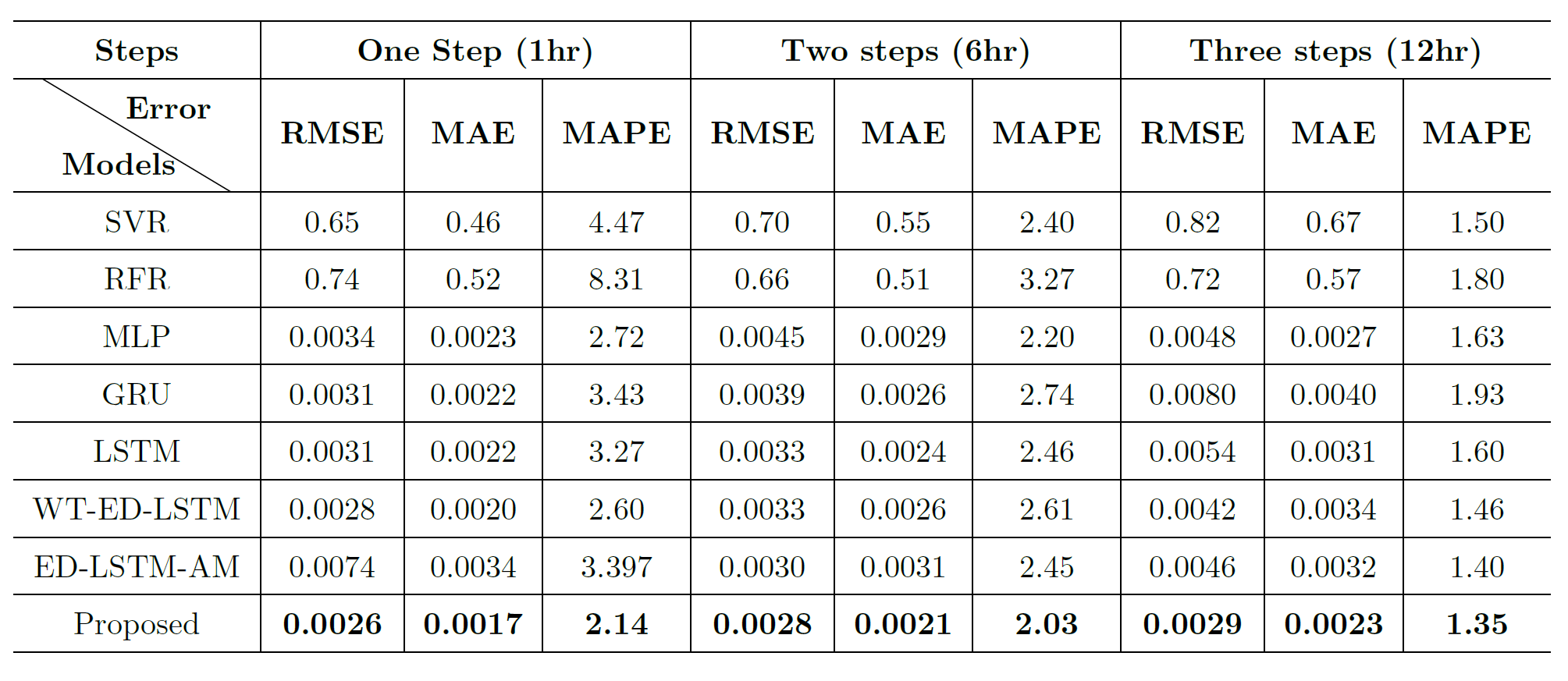}
\centering
\label{figtab1}
\end{figure}

\begin{figure}[tph!]
\includegraphics[width=1\textwidth,totalheight=5cm]{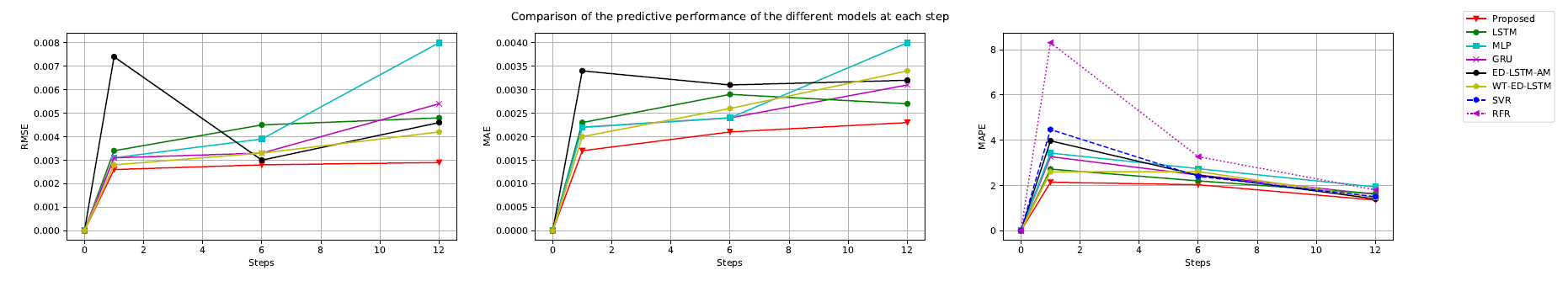}
\centering
\caption{Comparison of the predictive performance of the different models at each step.}
\label{fig4}
\end{figure}

Let us first discuss the one step prediction results.
The performance of the LSTM and GRU models for one-step-ahead prediction
were very similar, with the LSTM model showing an (edge) improvement
over GRU one, as far the MAPE criterion was concerned. The MLP model performance
was lower than LSTM and GRU when considering RMSE and MAE
criteria; it scored better than the LSTM and GRU when MAPE criterion was
considered.
The proposed approach (WT-ED-LSTM-AM) outperformed all baseline models on all multi-step prediction tasks. Figure \ref{fig4} illustrates this achievement, over
prediction steps ranging from 1 to 12.

\begin{figure}[tph!]
\includegraphics[totalheight=5cm]{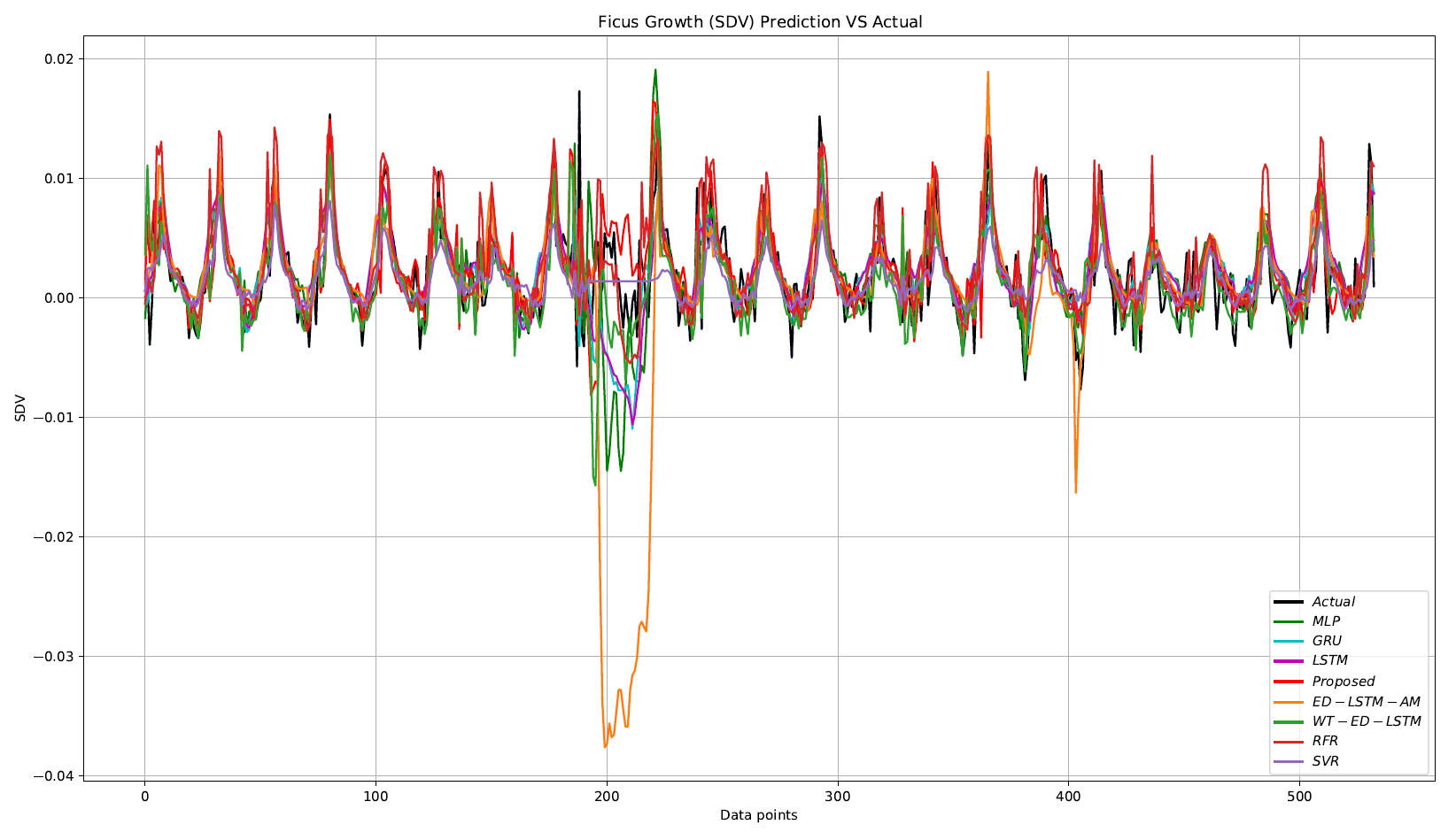}
\centering
\caption{Obtained accuracy in one-step prediction of Ficus growth (SVD) by the proposed
and baseline methods.}
\label{fig5}
\end{figure}

\begin{figure}[tph!]
\includegraphics[totalheight=10cm]{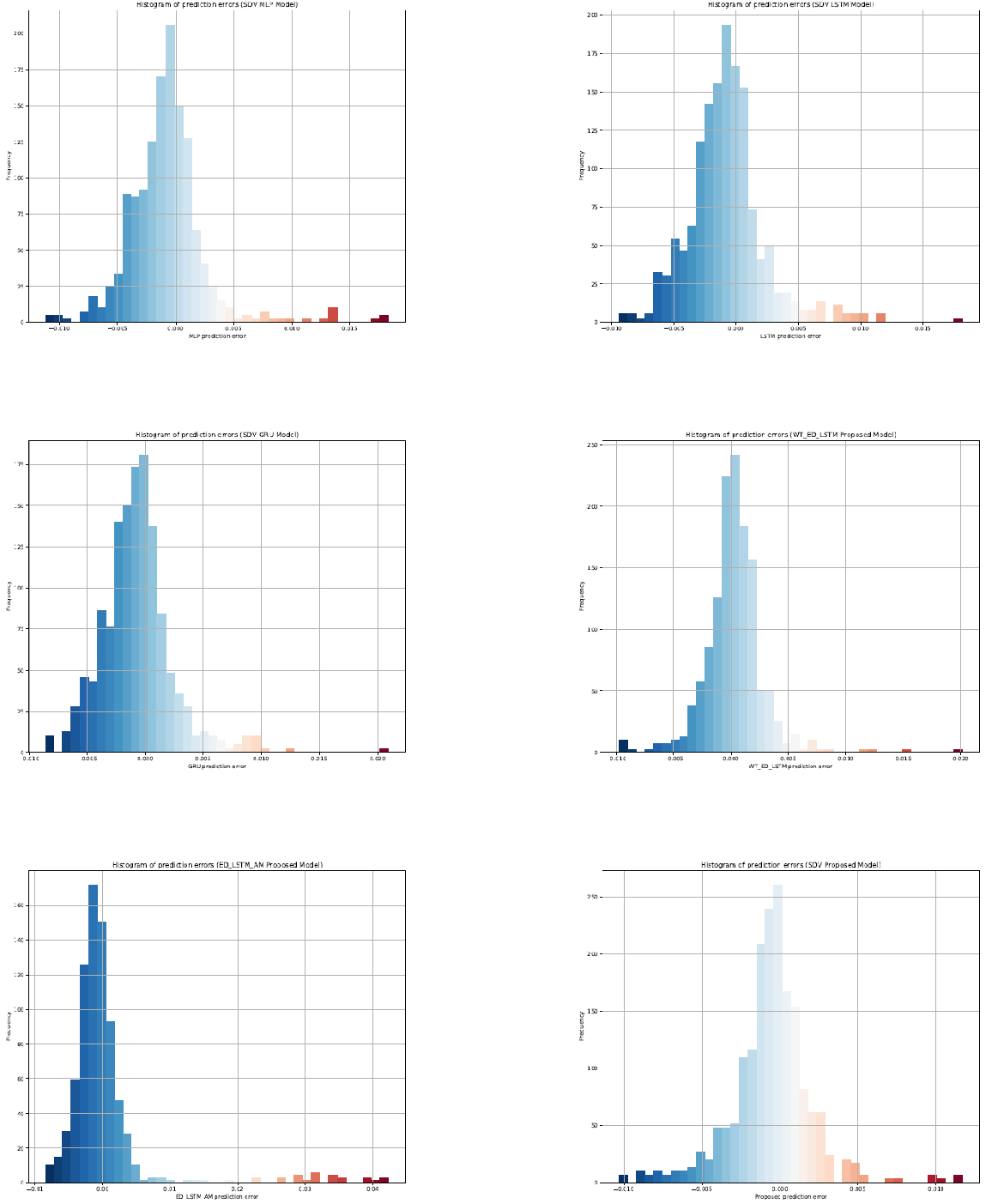}
\centering
\caption{Error distribution for one step prediction (1 hour ahead).}
\label{fig6}
\end{figure}

Fig. \ref{fig5} shows the accuracy of Ficus growth one-step prediction by all methods
for about 600 data samples. It can be seen that the proposed model successfully
performs one-step ahead prediction, outperforming the other methods
and providing accurate estimates of almost all peak values in the original data.
In addition, to compare the distribution of the prediction errors provided
by the baseline models with that of the proposed approach, we performed a
statistical analysis of them. The histograms of the produced one-step-prediction
errors are shown in Figure \ref{fig6}. It can be seen that in the proposed approach, close
to 57\% of predictions resulted in prediction errors around 0.00 and the remaining
43\% prediction errors ranged between -0.010 and 0.015.

Let us now discuss the results obtained in two-step prediction.
The proposed approach outperformed
all baseline models, providing lower RMSE, MAE and MAPE values in this case
as well.

\begin{figure}[tph!]
\includegraphics[totalheight=5cm]{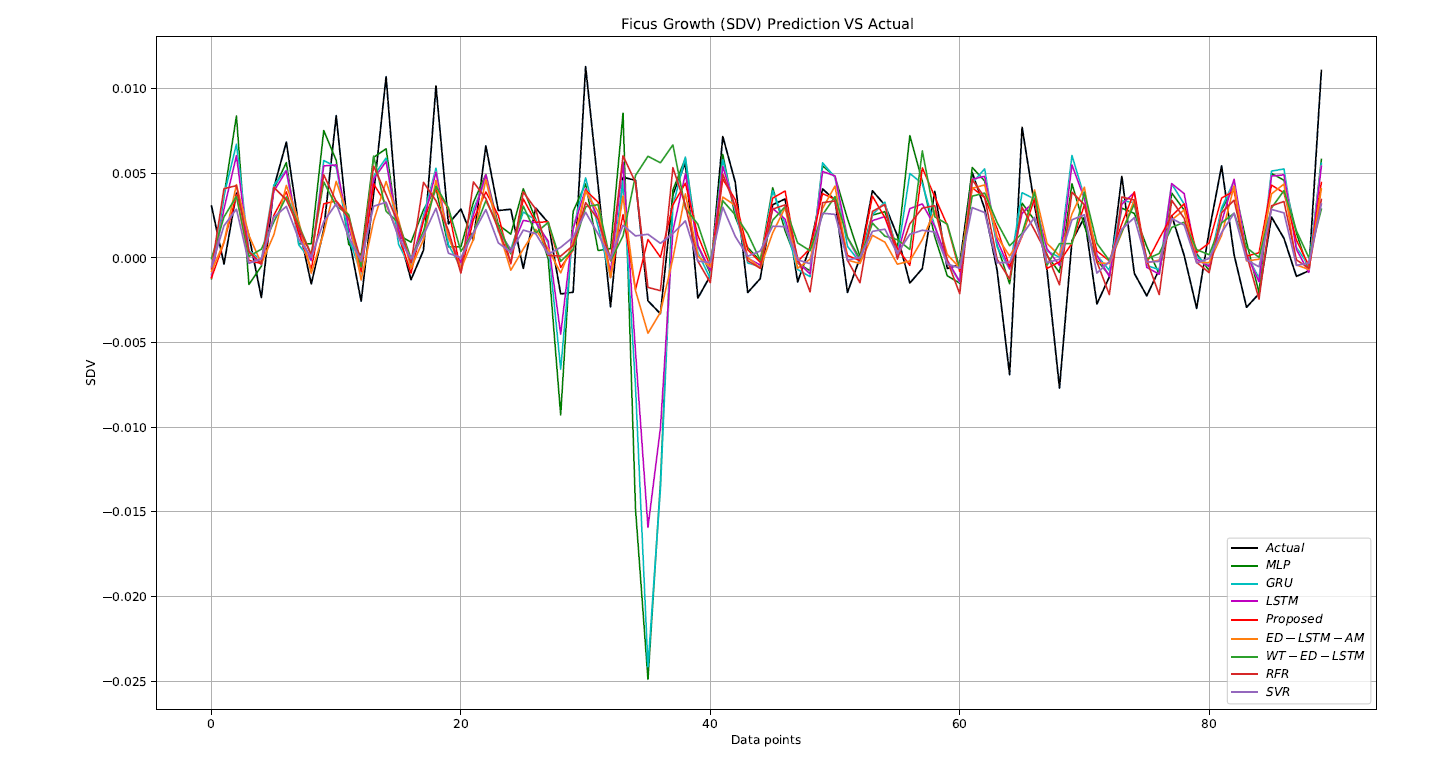}
\centering
\caption{Obtained accuracy in two-step prediction of Ficus growth (SVD) by the proposed
and baseline methods.}
\label{fig7}
\end{figure}

\begin{figure}[tph!]
\includegraphics[totalheight=10cm]{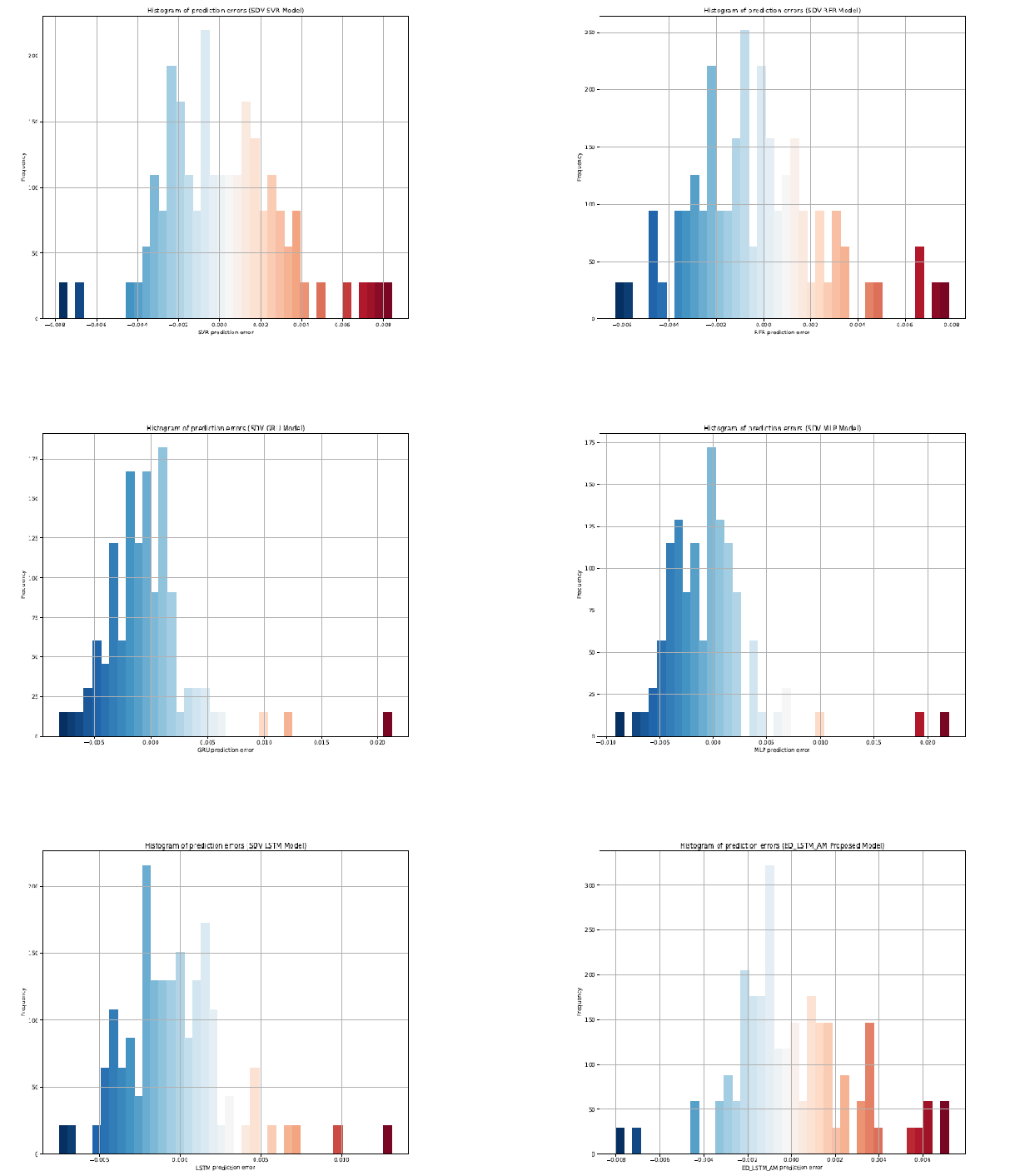}
\centering
\caption{Error distribution for two step prediction (6 hours ahead).}
\label{fig8}
\end{figure}

Fig. \ref{fig7} shows that almost all peak original values are precisely predicted by the proposed approach.
The resulting histograms for two-step-prediction errors are shown in Fig. \ref{fig8}.

Obtained accuracy in two-step prediction of Ficus growth (SVD) by the proposed
and baseline methods

In the proposed approach, close to 77\% of predictions resulted in prediction
errors between -0.004 and 0.002; the remaining 23\% of prediction errors ranged
between -0.008 and 0.008. This greatly outperformed the other baseline models.

The proposed approach also provided a superior three-step-ahead prediction.

\begin{figure}[tph!]
\includegraphics[totalheight=5cm]{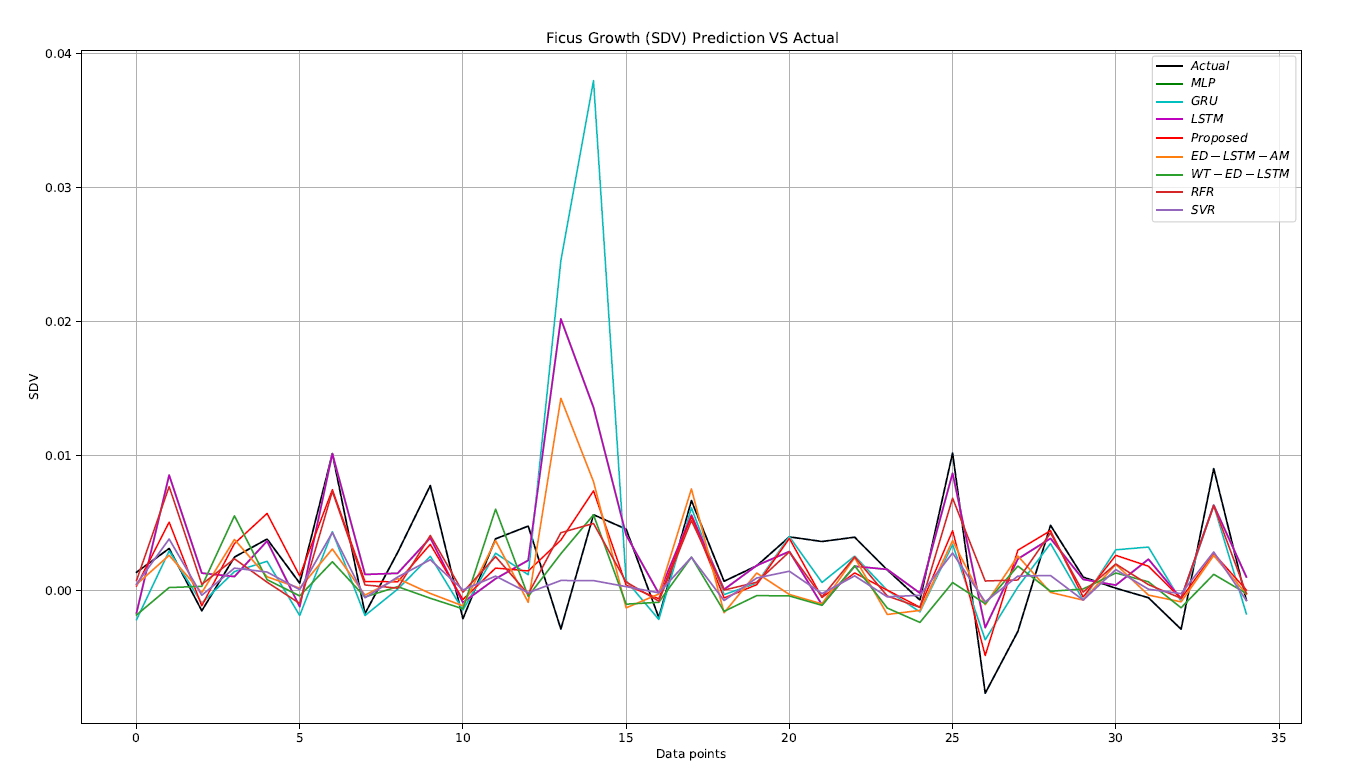}
\centering
\caption{Obtained accuracy in three-step prediction of Ficus growth (SVD) by the proposed
and baseline methods.}
\label{fig9}
\end{figure}

In Fig. \ref{fig9}, it can be seen that all baseline models failed to capture the peak
at data sample 14, with the proposed approach providing much better estimates
of the targeted values than the baseline models.

\begin{figure}[tph!]
\includegraphics[totalheight=10cm]{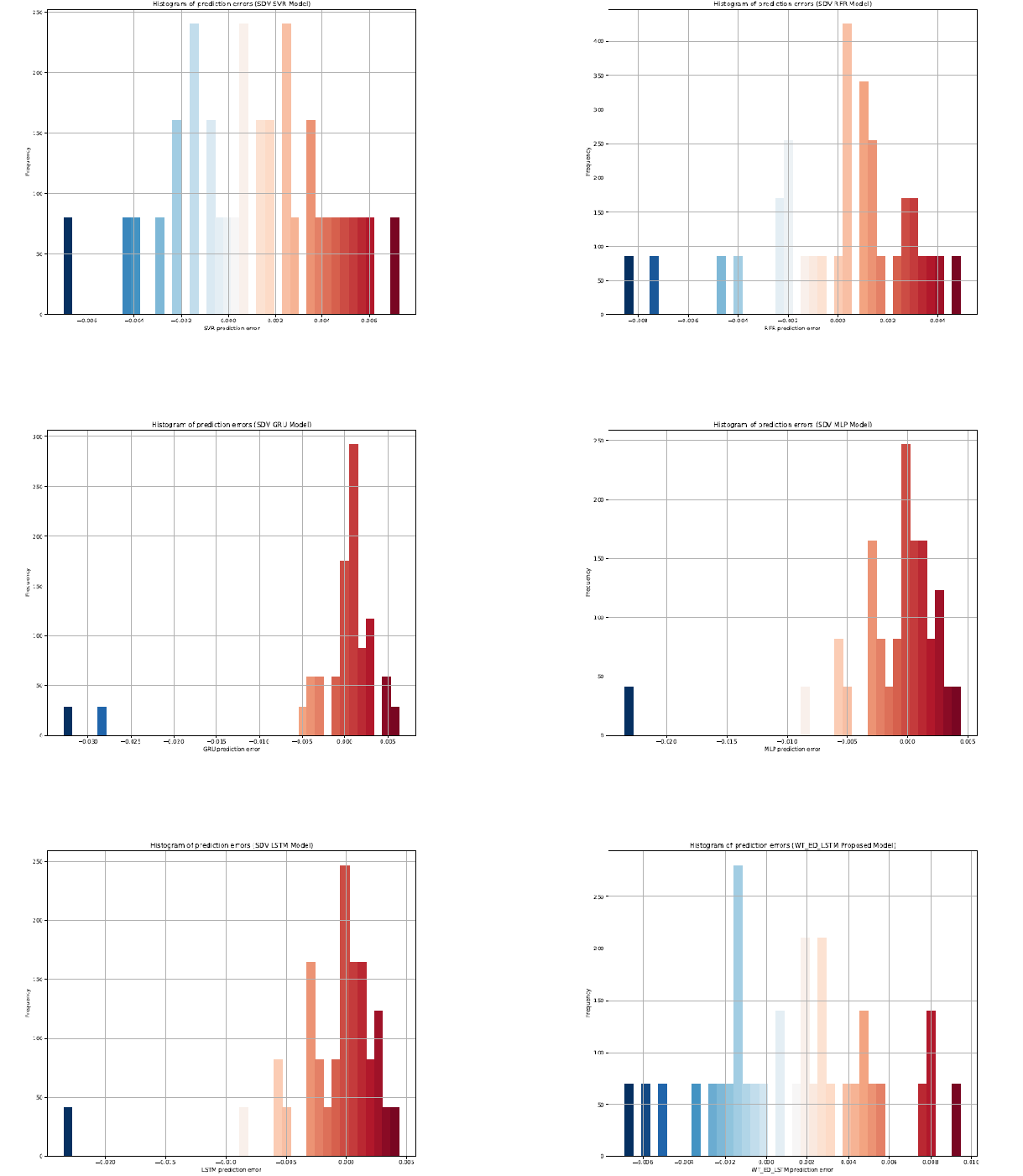}
\centering
\caption{Error distribution for three step prediction (12 hours ahead).}
\label{fig10}
\end{figure}

Fig. \ref{fig10} shows the prediction
error distributions for all baseline models, as well for the proposed approach.
It shows that close to 56\% of predictions provided by the proposed approach,
produced prediction errors between -0.002 and 0.002; the remaining 44\% of
prediction errors ranged between -0.006 and 0.006. In this case, as well, the
proposed method outperformed all other baseline models.

\section{Conclusions and Future Work} \label{s6}

This paper proposed a novel multi-step-ahead time series prediction approach.
The first step of the proposed method has been to use a wavelet
transform to decompose and smooth the original data. As a consequence, a
better model fitting could be achieved on the reconstructed signals. The second
step introduced an encoder-decoder framework based on LSTMs, which managed
to effectively produce appropriate features for multi-step prediction. The
third step which used LSTMs coupled with an attention mechanism was able to
successfully implement the prediction tasks.
The proposed approach was used for multi-step-ahead prediction of Ficus
Benjamina stem diameter variations, providing high prediction accuracy.
Real-world data have been collected and formed datasets that were used to
evaluate the proposed methodology. Hourly time intervals were used in the input
data, as well in our multi-step-ahead predictions. Comparisons were carried out,
over these real-world data, with state-of-the-art baseline models, showing that
the developed approach provides much better prediction results.

A topic of future research is to merge the data driven approach presented
in this paper with knowledge-based ones, especially for modelling the context,
i.e., the relations among the considered variables; we will be adapting former
research of ours in merging symbolic and subsymbolic approaches \cite{x20, x21, x22}.

\section{Acknowledgements}
This work was done in the framework and with the support of the EU Interreg
SMARTGREEN project (2017-2021). We would like to thank all growers
(UK and EU), for providing us with the presented datasets. We also wish
to thank the reviewers of the paper. Their valuable feedback, suggestions and
comments helped us to improve the quality of this work.


\bibliography{mybibfile}

\begin{thebibliography}{10}
\expandafter\ifx\csname url\endcsname\relax
  \def\url#1{\texttt{#1}}\fi
\expandafter\ifx\csname urlprefix\endcsname\relax\def\urlprefix{URL }\fi
\expandafter\ifx\csname href\endcsname\relax
  \def\href#1#2{#2} \def\path#1{#1}\fi

\bibitem{x0}
B.~Alhnaity, M.~Abbod, A new hybrid financial time series prediction model,
  Engineering Applications of Artificial Intelligence 95 (2020) 103873.

\bibitem{x2}
B.~Alhnaity, S.~Pearson, G.~Leontidis, S.~Kollias, Using deep learning to
  predict plant growth and yield in greenhouse environments, arXiv preprint
  arXiv:1907.00624.

\bibitem{x10}
J.~G. De~Gooijer, R.~J. Hyndman, 25 years of time series forecasting,
  International journal of forecasting 22~(3) (2006) 443--473.

\bibitem{x1}
O.~A. Arqub, Adaptation of reproducing kernel algorithm for solving fuzzy
  fredholm--volterra integrodifferential equations, Neural Computing and
  Applications 28~(7) (2017) 1591--1610.

\bibitem{x28}
J.~Schmidhuber, Deep learning in neural networks: An overview, Neural networks
  61 (2015) 85--117.

\bibitem{x23}
A.~Krizhevsky, I.~Sutskever, G.~E. Hinton, Imagenet classification with deep
  convolutional neural networks, Communications of the ACM 60~(6) (2017)
  84--90.

\bibitem{x3}
O.~A. Arqub, M.~Al-Smadi, S.~Momani, T.~Hayat, Application of reproducing
  kernel algorithm for solving second-order, two-point fuzzy boundary value
  problems, Soft Computing 21~(23) (2017) 7191--7206.

\bibitem{x34}
D.~Kollias, A.~Tagaris, S.~Stafylopatis, S.~Kollias, G.~Tagaris, Deep neural
  architectures for prediction in healthcare, Complex \& Intelligent Systems
  4~(2) (2018) 119--131.

\bibitem{x30}
A.~Sorjamaa, J.~Hao, N.~Reyhani, Y.~Ji, A.~Lendasse, Methodology for long-term
  prediction of time series, Neurocomputing 70~(16-18) (2007) 2861--2869.

\bibitem{x32}
S.~B. Taieb, G.~Bontempi, A.~F. Atiya, A.~Sorjamaa, A review and comparison of
  strategies for multi-step ahead time series forecasting based on the nn5
  forecasting competition, Expert systems with applications 39~(8) (2012)
  7067--7083.

\bibitem{x31}
A.~Sorjamaa, J.~Hao, N.~Reyhani, Y.~Ji, A.~Lendasse, Methodology for long-term
  prediction of time series, Neurocomputing 70~(16-18) (2007) 2861--2869.

\bibitem{x9}
J.~Daniel, P.-U. Andr{\'e}s, S.~H{\'e}ctor, B.~Miguel, T.~Marco, et~al., A
  survey of artificial neural network-based modeling in agroecology, in: Soft
  Computing applications in industry, Springer, 2008, pp. 247--269.

\bibitem{x26}
R.~Pouteau, J.-Y. Meyer, R.~Taputuarai, B.~Stoll, Support vector machines to
  map rare and endangered native plants in pacific islands forests, Ecological
  Informatics 9 (2012) 37--46.

\bibitem{x24}
K.~G. Liakos, P.~Busato, D.~Moshou, S.~Pearson, D.~Bochtis, Machine learning in
  agriculture: A review, Sensors 18~(8) (2018) 2674.

\bibitem{x29}
A.~Singh, B.~Ganapathysubramanian, A.~K. Singh, S.~Sarkar, Machine learning for
  high-throughput stress phenotyping in plants, Trends in plant science 21~(2)
  (2016) 110--124.

\bibitem{x14}
I.~Goodfellow, Nips 2016 tutorial: Generative adversarial networks, arXiv
  preprint arXiv:1701.00160.

\bibitem{x13}
M.~d. C.~A. Gonz{\'a}lez-Ch{\'a}vez, R.~Carrillo-Gonz{\'a}lez,
  A.~Cuellar-S{\'a}nchez, A.~Delgado-Alvarado, J.~Su{\'a}rez-Espinosa,
  E.~R{\'\i}os-Leal, F.~A. Sol{\'\i}s-Dom{\'\i}nguez, I.~E. Maldonado-Mendoza,
  Phytoremediation assisted by mycorrhizal fungi of a mexican defunct lead-acid
  battery recycling site, Science of the Total Environment 650 (2019)
  3134--3144.

\bibitem{x6}
A.~Chlingaryan, S.~Sukkarieh, B.~Whelan, Machine learning approaches for crop
  yield prediction and nitrogen status estimation in precision agriculture: A
  review, Computers and electronics in agriculture 151 (2018) 61--69.

\bibitem{x33}
M.~W. Vandegehuchte, A.~Guyot, M.~Hubau, S.~R. De~Groote, N.~J.
  De~Baerdemaeker, M.~Hayes, N.~Welti, C.~E. Lovelock, D.~A. Lockington,
  K.~Steppe, Long-term versus daily stem diameter variation in co-occurring
  mangrove species: Environmental versus ecophysiological drivers, Agricultural
  and Forest Meteorology 192 (2014) 51--58.

\bibitem{x19}
S.~Kanai, J.~Adu-Gymfi, K.~Lei, J.~Ito, K.~Ohkura, R.~E. Moghaieb, H.~El-Shemy,
  R.~Mohapatra, P.~K. Mohapatra, H.~Saneoka, et~al., N-deficiency damps out
  circadian rhythmic changes of stem diameter dynamics in tomato plant, Plant
  science 174~(2) (2008) 183--191.

\bibitem{x16}
T.~M. Hinckley, D.~N. Bruckerhoff, The effects of drought on water relations
  and stem shrinkage of quercus alba, Canadian Journal of Botany 53~(1) (1975)
  62--72.

\bibitem{x11}
L.~Duchesne, D.~Houle, Modelling day-to-day stem diameter variation and annual
  growth of balsam fir (abies balsamea (l.) mill.) from daily climate, Forest
  Ecology and Management 262~(5) (2011) 863--872.

\bibitem{x25}
S.~G. Mallat, A theory for multiresolution signal decomposition: the wavelet
  representation, IEEE transactions on pattern analysis and machine
  intelligence 11~(7) (1989) 674--693.

\bibitem{x8}
C.~Cortes, V.~Vapnik, Support-vector networks, Machine learning 20~(3) (1995)
  273--297.

\bibitem{x17}
T.~K. Ho, The random subspace method for constructing decision forests, IEEE
  transactions on pattern analysis and machine intelligence 20~(8) (1998)
  832--844.

\bibitem{x15}
S.~Haykin, N.~Network, A comprehensive foundation, Neural networks 2~(2004)
  (2004) 41.

\bibitem{x18}
S.~Hochreiter, J.~Schmidhuber, Long short-term memory, Neural computation 9~(8)
  (1997) 1735--1780.

\bibitem{x7}
J.~Chung, C.~Gulcehre, K.~Cho, Y.~Bengio, Empirical evaluation of gated
  recurrent neural networks on sequence modeling, arXiv preprint
  arXiv:1412.3555.

\bibitem{srivastava2015unsupervised}
N.~Srivastava, E.~Mansimov, R.~Salakhudinov, Unsupervised learning of video
  representations using lstms, in: International conference on machine
  learning, 2015, pp. 843--852.

\bibitem{x4}
Y.~Avrithis, N.~Tsapatsoulis, S.~Kollias, Broadcast news parsing using visual
  cues: A robust face detection approach, in: 2000 IEEE International
  Conference on Multimedia and Expo. ICME2000. Proceedings. Latest Advances in
  the Fast Changing World of Multimedia (Cat. No. 00TH8532), Vol.~3, IEEE,
  2000, pp. 1469--1472.

\bibitem{x27}
K.~Rapantzikos, N.~Tsapatsoulis, Y.~Avrithis, S.~Kollias, Bottom-up
  spatiotemporal visual attention model for video analysis, IET Image
  Processing 1~(2) (2007) 237--248.

\bibitem{x12}
Z.~Geng, G.~Chen, Y.~Han, G.~Lu, F.~Li, Semantic relation extraction using
  sequential and tree-structured lstm with attention, Information Sciences 509
  (2020) 183--192.

\bibitem{x5}
D.~Bahdanau, K.~Cho, Y.~Bengio, Neural machine translation by jointly learning
  to align and translate, arXiv preprint arXiv:1409.0473.

\bibitem{x20}
B.~Glimm, Y.~Kazakov, I.~Kollia, G.~B. Stamou, Lower and upper bounds for
  sparql queries over owl ontologies., in: AAAI, Citeseer, 2015, pp. 109--115.

\bibitem{x21}
I.~Kollia, B.~Glimm, I.~Horrocks, Answering queries over owl ontologies with
  sparql., in: OWLED, 2011.

\bibitem{x22}
D.~Kollias, G.~Marandianos, A.~Raouzaiou, A.-G. Stafylopatis, Interweaving deep
  learning and semantic techniques for emotion analysis in human-machine
  interaction, in: 2015 10th International Workshop on Semantic and Social
  Media Adaptation and Personalization (SMAP), IEEE, 2015, pp. 1--6.

\end{thebibliography}

\end{document}